\documentclass{ecai} 

\usepackage{latexsym}
\usepackage{amssymb}
\usepackage{amsmath}
\usepackage{amsthm}
\usepackage{booktabs}
\usepackage{enumitem}
\usepackage{graphicx}
\usepackage{color}
\usepackage{stfloats}
\usepackage{todonotes} 

\newcommand{\BibTeX}{B\kern-.05em{\sc i\kern-.025em b}\kern-.08em\TeX}

\begin{document}


\begin{frontmatter}


\paperid{123} 


\title{CoRA: Optimizing Low-Rank Adaptation with Common Subspace of Large Language Models}

\author[A]{\fnms{Xiaojun}~\snm{Xiao}}
\author[A]{\fnms{Sen}~\snm{Shen}}
\author[A]{\fnms{Qiming}~\snm{Bao}} 
\author[A]{\fnms{Hongfei}~\snm{Rong}}
\author[A]{\fnms{Kairui}~\snm{Liu}}
\author[A]{\fnms{Zhongsheng}~\snm{Wang}}
\author[A]{\fnms{Jiamou}~\snm{Liu}\thanks{Corresponding Author. Email: jiamou.liu@auckland.ac.nz.}}

\address[A]{The University of Auckland\\Auckland, Auckland}


\begin{abstract}
    In fine-tuning large language models (LLMs), conserving computational resources while maintaining effectiveness and improving outcomes within the same computational constraints is crucial. The Low-Rank Adaptation (LoRA) strategy balances efficiency and performance in fine-tuning large models by reducing the number of trainable parameters and computational costs. However, current advancements in LoRA might be focused on its fine-tuning methodologies, with not as much exploration as might be expected into further compression of LoRA. Since most of LoRA's parameters might still be superfluous, this may lead to unnecessary wastage of computational resources. In this paper, we propose \textbf{CoRA}: leveraging shared knowledge to optimize LoRA training by substituting its matrix $B$ with a common subspace from large models. Our two-fold method includes (1) Freezing the substitute matrix $B$ to halve parameters while training matrix $A$ for specific tasks and (2) Using the substitute matrix $B$ as an enhanced initial state for the original matrix $B$, achieving improved results with the same parameters. Our experiments show that the first approach achieves the same efficacy as the original LoRA fine-tuning while being more efficient than halving parameters. At the same time, the second approach has some improvements compared to LoRA's original fine-tuning performance. They generally attest to the effectiveness of our work.
\end{abstract}

\end{frontmatter}



\section{Introduction}

Reducing the parameter count necessary for model training could decrease computational expenses in maintaining equivalent training outcomes. Furthermore, optimizing the volume of parameters to achieve the same training performance is a critical objective, ensuring more efficient resource utilization. In an era characterized by numerous demands for training large language models (LLMs), cost reduction and performance enhancement are imperative for developing more advanced, sustainable, and cost-effective models. This approach could optimize resource utilization and broaden the opportunity for a wider range of individuals to train and deploy domain-specific large models in their respective fields, thereby overcoming the barrier of insufficient training resources that often hinder the development of targeted models.

\begin{figure}[!htbp]
  \centering
  \includegraphics[width=\linewidth]{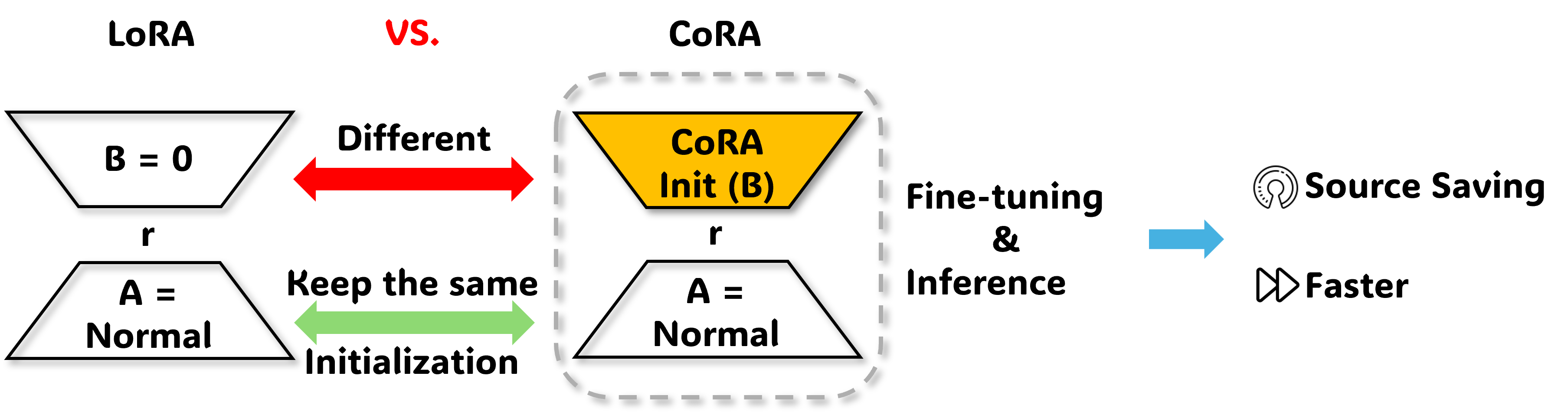}
  \caption{A simple case better explains the problems this paper focuses on and what CoRA is doing. A new method initializes the B matrix to replace the initialization in LoRA and complete the fine-tuning tasks.}
  \label{fig:intro}
\end{figure}

In the field of Natural Language Processing (NLP), the adaptation of large language models (LLMs) for downstream tasks employs various Parameter-Efficient Fine-Tuning (PEFT) methods \cite{pu2023empirical} to reduce training costs while maintaining or enhancing performance. Notable among these methods are techniques like Adapter Tuning \cite{houlsby2019parameter}. Adapter Tuning, exemplified by methods such as Low-Rank Adaptation (LoRA) \cite{hu2021lora}, involves introducing small, trainable Low-Rank matrices that adjust the model's existing weight matrices, thereby modifying the pre-trained model. This method allows the model to adapt quickly to new tasks while maintaining the stability of the pre-trained model structure. Adapter models achieve personalized adjustments while maintaining efficiency. These PEFT techniques substantially reduce the number of parameters that need training, enabling effective adaptation to new tasks without the computational cost of retraining the entire model. They lower the barriers to training large-scale models and expand their applicability across various tasks and domains.

However, the computational resources required for these methods remain substantial \cite{zhao2023survey}. We aim to reduce computational resource usage further and leverage existing resources to optimize large models for downstream tasks. Several methods have been developed to enhance the efficiency of LoRA, such as DyLoRA \cite{valipour2022dylora}, which prioritizes the representations learned by the adapter modules during training. Another method, QA-LoRA \cite{xu2023qa}, quantifies the weights of LLMs to reduce time and memory usage; after fine-tuning, LLM and auxiliary weights are seamlessly integrated into the quantized model without impacting accuracy. These strategies optimize LoRA across dimensions, focusing on learning efficiency and hardware adaptability. However, they maintain performance, they might not reduce the number of redundant parameters. Instead, these strategies improve the existing LoRA framework rather than fundamentally altering the LoRA architecture, which might still necessitate substantial GPU memory support for training many parameters. Therefore, we consider how to reduce parameters further or achieve favorable performance with the current parameter volume and how a universal method can be developed, allowing for broader application in future model training after complex processes. 

As demonstrated in Figure \ref{fig:intro}, we observed the B matrix within the LoRA structure and tried to optimize it. As we know, there are many parameters in each big model; there may be redundant parameters. So, our initial idea is to find valid parameters in the big model to replace the parameters of the B-matrix and reduce the training cost.

\begin{figure*}[!t]
\centering
\includegraphics[width=0.8\textwidth]{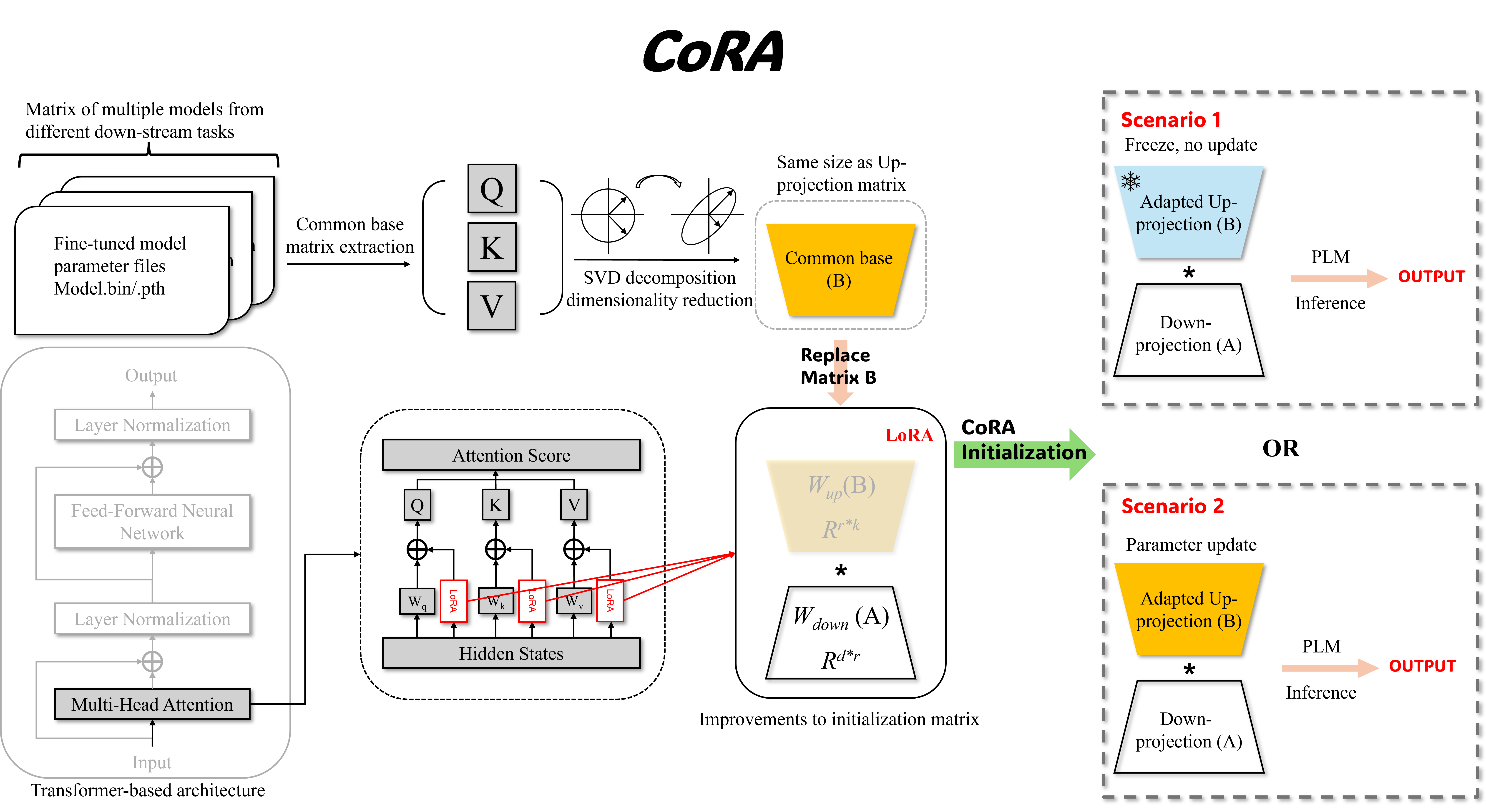}
\caption{An overview of CoRA. In the downstream large models, we extracted a common basis matrix within the corresponding attention heads' $Q$, $K$, and $V$ matrices. Utilizing Singular Value Decomposition (SVD) for dimensionality reduction, we adapted this matrix to meet the projection specifications required by the LoRA fine-tuning paradigm. The adapted common basis matrix replaced the original $B$ matrices in the LoRA on the corresponding attention heads' $Q$, $K$, and $V$ matrices. This modification was integrated into subsequent training processes. The fine-tuning was conducted using two methods: one where the $B$ matrix was replaced with a common basis matrix and then froze, and another where the $B$ matrix was replaced but kept training.}
\label{fig:parent}
\end{figure*}

Most current training systems extensively fine-tune pre-trained models for specific downstream tasks. We propose a hypothesis: these models keep core knowledge unchanged and have a common knowledge space inside each downstream model. After training, the base model activates knowledge in this space and provides domain-specific insights for specialized downstream applications. As Figure \ref{fig:parent} illustrated, we aim to extract this common subspace from multiple large models and employ a condensed module to replace the \(B\) matrix in LoRA training. The training process is delineated into two distinct scenarios. The first scenario involves replacing and then freezing the common basis matrix for the \(B\) matrix, thereby reducing the total parameter count of LoRA by 50\%, potentially eliminating several redundant parameters. This reduction could save computational resources and enhance training efficiency. The second scenario employs an alternative matrix as an optimized initialization for LoRA’s \(B\)  matrix, aiming to achieve favorable training outcomes within the same computational budget and ultimately performing generation comparable to traditional LoRA with a large model. Our contributions can be summarized as follows:

\begin{itemize}
\item We evaluate using a common basis matrix instead of the original matrix $B$ across various model scales. Our findings lend credence to the potential existence of a universal subspace within these models.

\item We freeze the common basis matrix after replacing it. This approach sometimes slightly surpasses the performance of traditional LoRA training methods while using only half the parameters, conserving computational resources.

\item We use the common basis matrix as an enhanced initial state for LoRA’s $B$ matrix. This strategy outperforms the original fine-tuning effectiveness of LoRA, reaching up to a 4\% peak improvement with favorable performance in fluency, relevance, and accuracy.
\end{itemize}

\begin{figure*}[!t]
\centering
\includegraphics[width=0.5\textwidth]{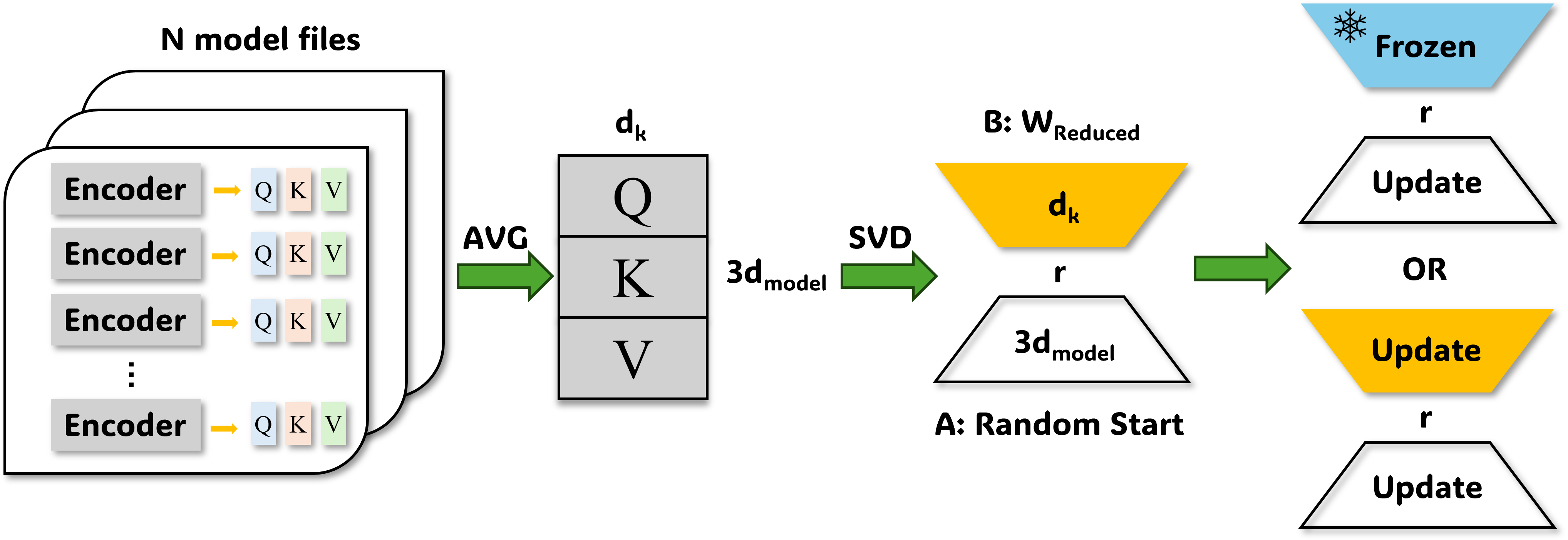}
\caption{Details in CoRA about extracting the common matrix space and performing SVD dimensionality reduction to adapt the $B$ matrix's expression form and apply it to lightweight large model fine-tuning.}
\label{fig:son}
\end{figure*}

\section{Related Work}

\subsection{Parameter-Efficient Fine-Tuning}

The exponential growth in the number of parameters in Transformer-based pre-trained language models (PLMs) \cite{vaswani2017attention}, particularly in large language models (LLMs) such as LlaMa \cite{touvron2023llama}, has led to significant advancements in a wide array of natural language processing (NLP) tasks. This surge in model complexity has substantially improved the effectiveness of these models, especially when they are fine-tuned for specific downstream datasets. Contrasting with zero-shot learning \cite{wei2021finetuned}, such fine-tuning further improves the models' capability to grasp and process task-specific subtleties, consequently elevating their overall performance.

However, the increasing scale of these models presents considerable challenges. The computational demands and resource requirements for fine-tuning these behemoths are substantial, especially when computational resources are limited. Addressing this issue, Parameter-Efficient Fine-Tuning (PEFT) methods \cite{ding2023parameter, liu2023gpt} have emerged as pivotal solutions. Techniques such as Prompt Tuning \cite{shin2020autoprompt, chen2022knowprompt} and Adapter Tuning \cite{houlsby2019parameter, newman2021p} represent innovative approaches in model fine-tuning. The former employs task-specific prompts for precise model adjustments, while the latter integrates a minimal number of specialized parameters for each task, effectively reducing overall computational demands. These methods achieve a balance, maintaining performance levels comparable to comprehensive fine-tuning while minimizing resource consumption. Nevertheless,  models may underperform when applied to new domains with unseen entity types. Prefix Tuning \cite{li2021prefix, vos2022towards} emerges as a viable solution, optimizing a small set of continuous, task-specific vectors as input, ensuring the effective processing of subsequent tokens. The development of these technologies provides feasible strategies for effectively utilizing advanced NLP large models in resource-constrained \cite{tabani2021improving} settings. 

Although all the aforementioned methods have markedly decreased the computational requirements, fine-tuning these extensive models still demands significant resources, especially in environments with limited computational capacity \cite{liao2023make}. Therefore, even parameter-efficient fine-tuning techniques can lead to prolonged computation times in low-resource settings such as those involving consumer-grade graphics cards.

\subsection{Low-Rank Adaption of LLMs}

Low-Rank Adaptation (LoRA) \cite{hu2021lora} represents a transformative approach in fine-tuning pre-trained deep learning models. This method enhances the model's attention mechanism by incorporating low-rank up-projection and down-projection matrices. It facilitates model adaptation to new tasks or datasets by targeting a minimal subset of parameters for adjustment, thereby maintaining the integrity of the original model's parameters. Its design prioritizes efficiency and flexibility, reducing training and inference times without compromising performance. However, a notable challenge with LoRA arises in scenarios with limited computational resources, such as edge computing devices, where it is difficult to balance model accuracy with computational and memory constraints.

Numerous variants of LoRA have been developed, each offering distinct enhancements to the fine-tuning process. One notable variant, QLoRA \cite{dettmers2023qlora, wang2023aligning}, employs 4-bit quantized backpropagation combined with Low-Rank Adapters to facilitate the efficient training of large models, accommodating up to 65 billion parameters on a single 48GB GPU. This variant introduces several technological advancements, such as the 4-bit NormalFloat data type, Double Quantization, and Paged Optimizers, designed to optimize memory utilization without degrading model performance. Another innovative variant, QA-LoRA \cite{xu2023qa}, builds upon the LoRA framework by incorporating quantization-aware techniques that enhance memory and computational efficiency during fine-tuning. It introduces group-wise operators for refined quantization and adaptation, effectively merging the advantages of LoRA with significant improvements in computational efficiency, which is particularly beneficial in resource-limited environments. Furthermore, DyLoRA \cite{valipour2022dylora} tackles the rigidity of traditional LoRA approaches by enabling training across a spectrum of ranks. This flexibility accelerates the training process 4 to 7 times, depending on the specific task, and maintains high performance across diverse pre-trained models like RoBERTa \cite{liu2019roberta}, and GPT \cite{floridi2020gpt}.

In contrast to existing research, which primarily aims to refine LoRA's architecture \cite{xu2023parameter} to boost inference efficiency and performance, our approach takes a fundamentally different direction. We identify several limitations inherent in previous implementations of LoRA, particularly the balance between model complexity and computational efficiency in resource-limited settings. Our method restructures the foundational elements of LoRA, specifically focusing on dramatically reducing the number of trainable parameters. This novel approach ensures that even in environments with strict resource limitations, the model remains functional and efficient, providing a viable solution to the challenges posed by the high resource demands of traditional fine-tuning methods.


\begin{figure*}[!t] 
  \centering 
  \includegraphics[width=\textwidth]{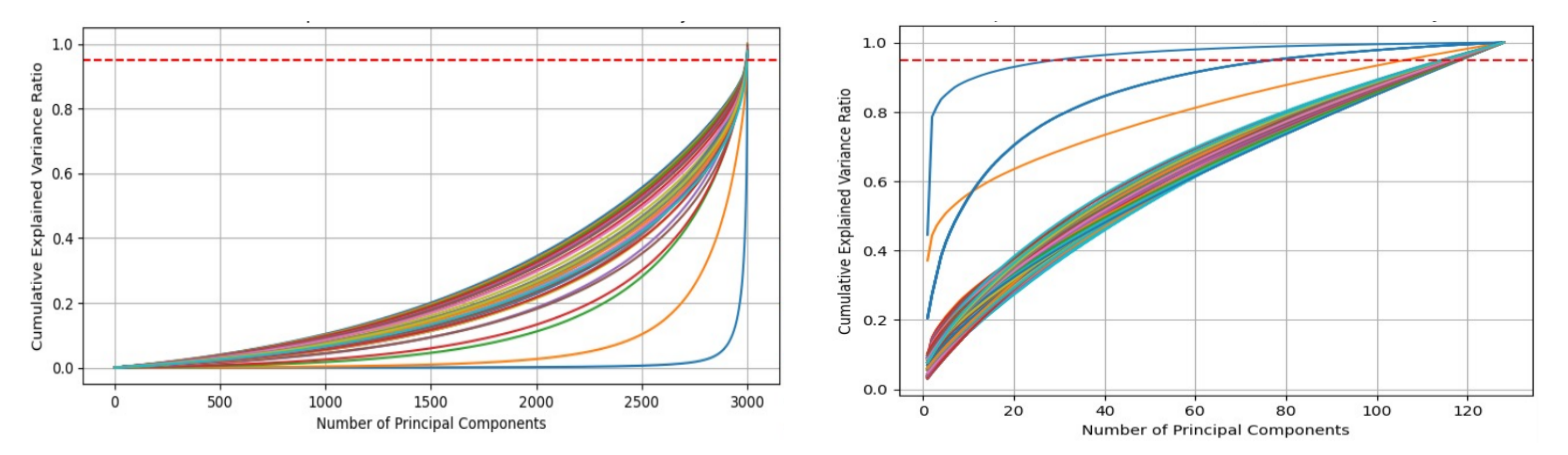} %

\caption{The number of principal components or singular values required to explain the variance using PCA (left) and SVD(right) techniques fully. PCA requires about 3000 principal components, while SVD only requires about 130 singular values, showing the efficiency of SVD in dimensionality reduction of high-dimensional data.}
\label{fig:SVD vs. PCA}
\end{figure*}

\section{Method}


Our experiment aimed to illustrate that by replacing and freezing the $B$ matrix in LoRA with the common basis matrix we extracted, the performance could be comparable to that achieved by fine-tuning the original LoRA parameters, thereby surpassing the efficiency of halving the LoRA parameters. Moreover, employing our extracted common basis matrix as an improved $B$ matrix initialization yielded training effectiveness that exceeded the original fine-tuning outcomes of LoRA.

\subsection{Workflow of The Entire Framework}
More details of CoRA are shown in Figure \ref{fig:son}. The entire experiment is based on Llama2-13B, including the complete attention head mechanism suitable for our related experiments. After downstream tasks fine-tuning, we can obtain each layer's $Q$, $K$, and $V$ matrices \cite{vaswani2017attention} from different models and put all of them into one matrix $W_{0} = [W_Q, W_K, W_V]^{T} \in \mathbb{R}^{3d_{model} \times d_k}$. Referring to the technical principles of LoRA, to effectively adjust the pre-trained Transformer model, LoRA targets the critical weight matrices $Q$, $K$, and $V$. These matrices usually contain many parameters, and LoRA combines them into two matrices with lower rankings (denoted $A \in \mathbb{R}^{3d_{model} \times r}$ and $B \in \mathbb{R}^{r \times d_{k}}$, and $r \ll min(d_{model}, d_k)$) to reduce the number of parameters. 

\begin{equation}
W = W_0 + AB, \quad \text{where } W_0 = [W_Q, W_K, W_V]^T
\end{equation}

$r$ is the target representation dimension of the word vector in our LoRA dimensionality reduction and dimensionality process. This low-rank approximation reduces the parameters' scale and significantly lowers computational and storage demands. During the fine-tuning phase of the model, instead of directly adjusting the original $Q$, $K$, and $V$ matrices, we focus on adjusting these two low-rank matrices, $A$ and $B$. By adjusting fewer parameters, LoRA effectively maintains the expressive power of the model while enhancing the efficiency of parameter adjustments. We chose to replace matrix $B$ because it is initialized to zero to ensure that the initial impact of the low-rank update is neutral at the start of training, meaning LoRA has minimal impact on the original pre-trained model at this stage. This approach maintains the stability of the pre-trained weights at the start of the fine-tuning process. It makes the fine-tuning transition smoother, allowing the model to adapt to new tasks while retaining the knowledge from pretraining. Matrix $A$ is randomly initialized with a normal distribution \cite{wang2023lora}. It provides an initial, non-zero gradient so the model can adjust these values during training to better adapt to new tasks. Therefore, we decided to replace $B$ and not $A$, as $A$ needs to update weights to better adapt to downstream tasks. The public space matrix $B$ originates from an in-depth analysis of various downstream tasks that were fine-tuned using the same dataset. Initially, we extracted the $Q$, $K$, and $V$ matrices from these $n$ models, which had been adapted to different tasks. We integrated these matrices into a single matrix $W_0$ by averaging them. 

\begin{equation}
W_0 = \frac{1}{n}\sum_{i=1}^nW_{i} = [W_Q{}_i, W_K{}_i, W_V{}_i]^{T}
\end{equation}


We employ $r$ as the target dimension in our matrix dimensionality reduction process, adapting the public space matrix to the elevated dimension matrix $B$. This matrix was subjected to Singular Value Decomposition (SVD) to reduce dimensionality, preserving only the top $r$ dimensions. These dimensions form the foundation of our public space matrix $B$, which acts as a dimensional adaptor within the LoRA framework. This setup enhances efficient parameter reduction while ensuring the model remains adaptable across various tasks. The following section will discuss the experimental basis for selecting this strategy.

\begin{equation}
W_0 = U \Sigma V^T
\label{eq:svd3}
\end{equation}

\begin{equation}
so\qquad V^T = \Sigma^{-1}U^TW_0    
\label{eq:svd4}
\end{equation}

We obtained the three data variables ($U \in \mathbb{R}^{3d_{model} \times 3d_{model}}, \Sigma \in \mathbb{R}^{3d_{model} \times d_{k}}, V^T \in \mathbb{R}^{d_{k} \times d_{k}}$) on the right through $W_0$ decomposition, where $V^T$ is the transpose of the right singular vector matrix $V$, and we take the first $r$ as the dimensionality reduction representation of the common space matrix.

\begin{equation}
W_{reduced} = B = V^T[:r, :]
\label{eq:svd5}
\end{equation}

\begin{equation}
W = W_0 + AW_{Reduced}
\end{equation}

As shown in the subgraph, after replacing matrix $B$, we adopted two different approaches: 1) We froze matrix $B$ because the replaced $B$ matrix, derived from the model adapted to downstream tasks, is assumed to be a well-trained matrix that can adapt to various downstream tasks, acting as a well-performing parameter matrix for these tasks. As mentioned, we need matrix $A$ to capture information relevant to new tasks, so we involved matrix $A$ in the training of the model and the iteration of parameters to improve performance in downstream tasks; 2) We involved matrix $B$ in parameter iteration because we wanted to test whether it could further adapt to downstream tasks during training, alongside matrix $A$, to fit better and improve the model's overall performance in these tasks.

\subsection{Methods For Matrix Compression Extraction}

In feature extraction and data dimensionality reduction, Principal Component Analysis (PCA) \cite{mackiewicz1993principal} and Singular Value Decomposition (SVD) \cite{abdi2007singular} are pivotal technologies. A thorough analysis of these methodologies provides insight into their application strengths and delineates their performance boundaries.

PCA identifies directions of maximum variance in the data, termed principal components, by eigendecomposition of the covariance matrix $\Sigma$, where the covariance matrix is given by: 

\begin{equation}
\Sigma = \frac{1}{3d_{model}-1}\sum^{n}_{i=1}(W_0 - \mu)(W_0- \mu)^T
\end{equation}

\begin{equation}
\Sigma = PDP^{-1}
\end{equation}

\begin{equation}
W'_{reduced} = W_0 \cdot P[:, 1:r]
\end{equation}


Here, \(W_0\) represents the matrix formed by merging the \(Q\), \(K\), and \(V\) matrices, and \(\mu\) denotes the mean of all data points in \(W_0\). The principal components are the eigenvectors of the covariance matrix \(\Sigma\), arranged in descending order according to their corresponding eigenvalues. The matrix \(P\) contains these eigenvectors, with each column representing an eigenvector. We select the first \(r\) eigenvectors from \(P\). This technique effectively simplifies the complexity of the data while preserving the majority of the information.

On the other hand, SVD, a more versatile decomposition method applicable to square and rectangular matrices, reveals the fundamental structure of data. During dimensionality reduction, SVD efficiently compresses data by retaining only a limited number of the largest singular values and their associated singular vectors. Formula (\ref{eq:svd3})(\ref{eq:svd4})(\ref{eq:svd5}) are the specific mathematical expressions of SVD in this matrix dimension reduction task. The following experiments confirmed that it was an effective part of our work and applied in our work. 

For the $Q$, $K$, $V$ matrices of all layers in the Transformer part of the Llama2 fine-tuned model architecture, we considered using SVD and PCA to reduce the dimensionality of the existing high-dimensional matrix data and observe changes in the real-time interpretation rate. The result can be seen in Figure \ref{fig:SVD vs. PCA}. Our analysis shows that achieving a 100\% explained variance rate—retaining all feature variability—requires approximately 3000 principal components with PCA, but SVD achieves a full variance explanation with just 130 singular values.

Thus, informed by these comparative results, we have meticulously chosen SVD as the dimensionality reduction technique for our CoRA lightweight fine-tuning strategy. This decision is predicated on SVD drastically reducing the components required while ensuring a complete variance explanation. This advantage is particularly significant when dealing with large or complex datasets, as it not only effectively compresses data but also reduces computational costs, thereby accelerating the model training and prediction process.

\subsection{Selection of Replacement Matrices}

\begin{table*}[!b]
\centering
\resizebox{0.6\textwidth}{!}{
\begin{tabular}{lcccccc}
\toprule
\textbf{Method} & \textbf{ROUGE-1} & \textbf{ROUGE-2} & \textbf{ROUGE-L} & \textbf{METEOR Score} & \textbf{SacreBLEU Score} & \textbf{BERT Score} \\ 
\midrule
CoRA\_yahma\_FB\_1 & 0.395 & 0.211 & 0.372 & 0.301 & 15.311 & 0.892 \\ 
CoRA\_yahma\_FB\_2 & 0.397 & 0.211 & 0.373 & 0.306 & 15.957 & 0.892 \\ 
CoRA\_yahma\_FB\_3 & 0.395 & 0.212 & 0.373 & 0.306 & 16.007 & 0.891 \\ 
CoRA\_yahma\_FB\_4 & 0.397 & 0.213 & 0.375 & 0.306 & 15.707 & 0.892 \\ 
CoRA\_yahma\_FB\_5 & 0.401 & 0.217 & 0.380 & 0.307 & 15.631 & 0.893 \\ \hline
CoRA\_yahma\_TB\_1 & 0.419 & 0.231 & 0.394 & 0.322 & 17.184 & 0.898 \\ 
CoRA\_yahma\_TB\_2 & 0.418 & 0.230 & 0.395 & 0.319 & 16.943 & 0.897 \\ 
CoRA\_yahma\_TB\_3 & 0.424 & 0.231 & 0.399 & 0.321 & 17.093 & 0.898 \\ 
CoRA\_yahma\_TB\_4 & 0.424 & 0.231 & 0.399 & 0.322 & 17.063 & 0.898 \\ 
CoRA\_yahma\_TB\_5 & 0.421 & 0.230 & 0.396 & 0.320 & 17.296 & 0.897 \\ \hline
CoRA\_code\_FB\_1 & 0.391 & 0.226 & 0.380 & 0.302 & 25.688 & 0.888 \\ 
CoRA\_code\_FB\_2 & 0.393 & 0.227 & 0.382 & 0.302 & 25.578 & 0.888 \\ 
CoRA\_code\_FB\_3 & 0.395 & 0.230 & 0.384 & 0.304 & 25.499 & 0.886 \\ 
CoRA\_code\_FB\_4 & 0.393 & 0.229 & 0.383 & 0.303 & 25.757 & 0.888 \\ 
CoRA\_code\_FB\_5 & 0.400 & 0.233 & 0.389 & 0.306 & 26.053 & 0.889 \\ \hline
CoRA\_code\_TB\_1 & 0.414 & 0.252 & 0.404 & 0.327 & 27.692 & 0.894 \\ 
CoRA\_code\_TB\_2 & 0.413 & 0.248 & 0.403 & 0.326 & 27.461 & 0.894 \\ 
CoRA\_code\_TB\_3 & 0.413 & 0.250 & 0.403 & 0.326 & 27.814 & 0.894 \\ 
CoRA\_code\_TB\_4 & 0.409 & 0.246 & 0.400 & 0.324 & 27.180 & 0.893 \\ 
CoRA\_code\_TB\_5 & 0.418 & 0.255 & 0.407 & 0.330 & 28.874 & 0.894 \\ \bottomrule
\end{tabular}
}
\caption{Performance evaluation of downstream tasks on the text-to-code dataset after fine-tuning with the common basis matrix extracted from varying numbers of models to replace the $B$ matrix. The table illustrates the inference results assessed by the evaluation method. [TB means that the replaced matrix continues training, and FB means that the replaced matrix remains frozen.]}
\label{SVD Result}
\end{table*}

To observe how many models we require to extract an exceptional matrix that could serve as our $B$ matrix, we randomly selected 1, 2, 3, 4, or all 5 models from a pool of 5 for our experiments. The random selection ensured fairness and randomness in the experiment, aiming for relatively unbiased results. The 5 models were chosen from different downstream task models to investigate the possibility of a common basis matrix that could be extracted and utilized.

From our experiments, as shown in the picture \ref{SVD Result}, it is evident that whether or not the $B$ matrix is updated, the performance remains stable across the yahma \cite{ji2024text} dataset and our text-to-code dataset. This stability may support several inferences: 1) Even after fine-tuning to adapt to downstream tasks, large models retain a common knowledge space—the common basis matrix referred to in this paper—which leads to stable results regardless of the number of matrices extracted; 2) Updating the $B$ matrix does seem to yield better results than training the $A$ matrix alone, although this could be due to the effect of updating a larger number of parameters. Concrete conclusions will require further experiments to prove that better performance can be achieved with the same resources after a suitable initialization of the $B$ matrix.

Through these experiments, we conjecture that any of the matrices extracted could be used for subsequent experiments due to the presence of a common basis matrix. However, there may be unknown factors causing some matrices to be extracted less optimally. For the sake of accuracy in our results, we chose the matrix extracted from the 5 models that showed the best and most stable combined performance for our subsequent experiments.


\section{Experimental Settings}
We conducted the entirety of our experiment using the Llama2 13B-hf \cite{touvron2023llama} as our frozen large language model of choice.

\textbf{Dataset:} We used 2 datasets in our experiments: yahma \cite{alpaca} from Hugging Face and a customed text-to-code dataset, to train downstream tasks. These instruction data can be used to perform instruction tuning on the language model to make it follow instructions better. The collation datasets are performed by collating the incoming GPT-4 for better model performance.

\textbf{Evaluation:} We used a comprehensive set of evaluation metrics to validate our experiment results robustly. These included ROUGE-1 \cite{lin2004rouge}, ROUGE-2, and ROUGE-L for assessing text similarity at different levels—individual words, bi-grams, and sentence structure. Additionally, we incorporated the Meteor Score \cite{denkowski2014meteor} for its nuanced evaluation of translation quality through precision, recall, and synonym matching. The SacreBLEU Score \cite{post2018call} provided a standardized approach to machine translation evaluation, while the BERT Score \cite{zhang2019bertscore} evaluated semantic similarity by computing the cosine similarity between word vectors of the candidate and reference texts.

\textbf{LoRA Setting:} The foundational PEFT strategy employed in our experiment was LoRA, utilizing it from the PEFT library as a benchmark. We conducted experiments with \(r=48\), and for comparative purposes against the freezing strategy, we trained a baseline LoRA at \(r=24\) as well. 

\textbf{GPU:} The training process of our proposed method was carried out on NVIDIA® A100 GPUs with 80GB of memory.

\textbf{Model:} For extracting the common basis matrix, we randomly selected five models from Hugging Face. To ensure the fairness of the experiment, control experiments were conducted by randomly selecting and extracting the common matrices from any one, two, three, four, or all five models for comparison.


\section{Result}

\begin{table*}[!t]
\centering
\resizebox{0.6\textwidth}{!}{
\begin{tabular}{lcccccc}
\toprule
\textbf{Method} & \textbf{ROUGE-1} & \textbf{ROUGE-2} & \textbf{ROUGE-L} & \textbf{METEOR Score} & \textbf{SacreBLEU Score} & \textbf{BERT Score} \\ 
\midrule
yahma\_FineTuning & 0.359 & 0.170 & 0.336 & 0.273 & 11.15871 & 0.887 \\
LoRA\_yahma\_48 & 0.397 & 0.214 & 0.374 & 0.293 & 14.587 & 0.893 \\
CoRA\_yahma\_FB\_5 & 0.401 & 0.217 & 0.380 & 0.307 & 15.631 & 0.893 \\ 
CoRA\_yahma\_TB\_5 & 0.421 & 0.230 & 0.396 & 0.320 & 17.296 & 0.897 \\ 
LoRA\_yahma\_24 & 0.389 & 0.205 & 0.367 & 0.294 & 14.709 & 0.891 \\ \hline
code\_FineTuning & 0.463 & 0.355 & 0.456 & 0.400 & 23.7442 & 0.885 \\
LoRA\_code\_48 & 0.393 & 0.231 & 0.381 & 0.300 & 25.758 & 0.889 \\
CoRA\_code\_FB\_5 & 0.400 & 0.233 & 0.389 & 0.306 & 26.053 & 0.889 \\ 
CoRA\_code\_TB\_5 & 0.418 & 0.255 & 0.407 & 0.330 & 28.874 & 0.894 \\ 
LoRA\_code\_24 & 0.384 & 0.220 & 0.373 & 0.302 & 25.499 & 0.889 \\ \bottomrule
\end{tabular}
}
\caption{\footnotesize A comparative analysis was conducted against traditional LoRA training methods with the same rank, traditional LoRA training methods with half the parameter quantity, and full parameter iterative training. Here, `48' refers to the same rank as FB\_5 and TB\_5, while `24' corresponds to the parameter quantity of the $B$ matrix after freezing, i.e., the same as FB\_5. Subsequently, the generated content was evaluated using various performance metrics.
}
\label{result table}
\end{table*}

\begin{table*}[!b]
\centering
\resizebox{0.6\textwidth}{!}{
\begin{tabular}{lccccc}
\toprule
\textbf{Method} & \textbf{ROUGE-1} & \textbf{ROUGE-2} & \textbf{ROUGE-L} & \textbf{METEOR Score} & \textbf{SacreBLEU Score}  \\
\midrule
LoRA\_code\_8 & 0.401 & 0.232 & 0.387 & 0.313 & 25.558  \\
CoRA\_code\_TB\_8 & 0.420 & 0.256 & 0.410 & 0.335 & 27.830 \\
CoRA\_code\_FBL\_8 & 0.414 & 0.250 & 0.404 & 0.333 & 27.859  \\
\hline
LoRA\_code\_16 & 0.397 & 0.231 & 0.386 & 0.310 & 26.228  \\
CoRA\_code\_TB\_16 & 0.423 & 0.259 & 0.412 & 0.343 & 28.757  \\
CoRA\_code\_FB\_16 & 0.418 & 0.255 & 0.408 & 0.332 & 27.502  \\
\hline
LoRA\_code\_32 & 0.396 & 0.229 & 0.385 & 0.306 & 26.248  \\
CoRA\_code\_TB\_32 & 0.423 & 0.258 & 0.412 & 0.340 & 29.045  \\
CoRA\_code\_FB\_32 & 0.420 & 0.256 & 0.409 & 0.336 & 27.981  \\
\hline
LoRA\_yhama\_8 & 0.396 & 0.213 & 0.372 & 0.314 & -  \\
CoRA\_yahma\_TB\_8 & 0.430 & 0.240 & 0.406 & 0.329 & 16.764  \\
CoRA\_yahma\_FB\_8 & 0.429 & 0.235 & 0.403 & 0.334 & 17.109  \\
\hline
LoRA\_yhama\_16 & 0.401 & 0.215 & 0.378 & 0.316 & 16.136  \\
CoRA\_yahma\_TB\_16 & 0.429 & 0.240 & 0.405 & 0.329 & 16.791  \\
CoRA\_yahma\_FB\_16 & 0.427 & 0.235 & 0.402 & 0.334 & 17.232  \\
\hline
LoRA\_yhama\_32 & 0.405 & 0.220 & 0.382 & 0.318 & 16.287  \\
CoRA\_yahma\_TB\_32 & 0.436 & 0.243 & 0.412 & 0.344 & 17.542  \\
CoRA\_yahma\_FB\_32 & 0.426 & 0.239 & 0.402 & 0.328 & 16.486  \\ 
\bottomrule
\end{tabular}
}
\caption{Overview of Experimental Results: Displaying model performance at various matrix ranks to validate the method's versatility. $TB/FB$ represents whether the B matrix in CoRA participates in the parameter update process. The number at the end is the dimension parameter of $r$.}
\label{Tongyong}
\end{table*}

The comparison results of LoRA fine-tuning and multiple versions of CoRA fine-tuning under the same rank are recorded in Table \ref{result table}, which indicate that the FB models (yahma\_FB\_5) marginally outperformed the LoRA (yahma\_48) on two datasets. In the yahma dataset, compared to the traditional FineTuning method, the remaining three models showed improvements across all evaluation metrics. Specifically, the TB\_5 model achieved scores of 0.421 in ROUGE-1, 0.230 in ROUGE-2, and 0.396 in ROUGE-L; its METEOR and SacreBLEU scores were 0.320 and 17.296, respectively, with a BERT score of 0.897, outperforming other methods.

The FB\_5 and TB\_5 models also demonstrated considerable advantages in evaluating the code-to-text dataset. The TB\_5 model scored 0.418 in ROUGE-1, 0.255 in ROUGE-2, and 0.407 in ROUGE-L; its METEOR score increased to 0.330, and its SacreBLEU and BERT scores reached 28.874 and 0.894, respectively. These results confirm its adaptability and efficiency in complex generation tasks.

The data validate the efficacy of utilizing a common basis matrix as an optimized initial state for the $B$ matrix in LoRA, which maintained performance with reduced computational resources and achieved favorable training outcomes within the same computational budget. Furthermore, these metrics also well confirm our original hypothesis that when trained in conjunction with the $A$ matrix, the common basis matrix can better retain the knowledge of the original pre-trained model and adapt to new downstream tasks.

\begin{table*}[!t]
\centering
\resizebox{0.6\textwidth}{!}{
\begin{tabular}{lcccccc}
\toprule
\textbf{Method} & \textbf{ROUGE-1} & \textbf{ROUGE-2} & \textbf{ROUGE-L} & \textbf{METEOR Score} & \textbf{SacreBLEU Score} & \textbf{BERT Score} \\
\midrule
yahma\_48\_0 & 0.243 & 0.129 & 0.232 & 0.168 & 2.296 & 0.853 \\
yahma\_48\_1 & 0.001 & 0.000 & 0.001 & 0.002 & 0.0005 & 0.710 \\
yahma\_48\_ran & 0.351 & 0.149 & 0.329 & 0.198 & 11.979 & 0.863 \\
\midrule
code\_48\_0 & 0.135 & 0.075 & 0.131 & 0.106 & 2.861 & 0.791 \\
code\_48\_1 & 0.0004 & 0.000 & 0.0004 & 0.004 & 0.0002 & 0.690 \\
code\_48\_ran & 0.363 & 0.131 & 0.341 & 0.252 & 21.874 & 0.869 \\
\bottomrule
\end{tabular}
}
\caption{Ablation Study: Assessing the Impact of the $B$ Matrix on Model Performance. This study includes experiments using a zero matrix for baseline evaluation, testing uniform non-zero values in the $B$ matrix to provide a constant bias, and examining whether performance enhancements with an extracted $B$ matrix can be replicated by a randomly initialized $B$ matrix, thereby assessing the unique contributions of our extracted matrix.}

\label{XiaoRong}
\end{table*}

Across different parameter settings ($r=8,\ 16,\ 32$), shown by Table \ref{Tongyong}, exhibited marginally better performance than traditional LoRA. Particularly in ROUGE-L and METEOR scores, the FB model exhibited better linguistic coherence and overall quality control, indicating the FB strategy's advantage in handling complex texts. Regarding SacreBLEU scores, the FB strategy also showed better stability and a slight lead, further validating its effectiveness in ensuring translation quality and alignment with human evaluation standards.

The TB-enhanced strategy showed improved results compared to traditional LoRA across various parameter settings. In measures such as ROUGE-1, ROUGE-2, and ROUGE-L scores, the TB model demonstrated effective understanding and reproducing details. Particularly at higher parameter settings (\(r=32\)), the TB strategy outperformed traditional LoRA in METEOR and SacreBLEU scores. These results underscore the usefulness of the FB and TB strategies in enhancing model performance and point to their potential in supporting the model's ability to handle complex linguistic tasks.

\section{Ablation Study}
We conducted the following ablation studies to determine if other factors could be influencing our experimental results or to identify potentially unstable variables. These studies are designed to ensure fairness and accuracy in our findings, resulting in the Figure \ref{XiaoRong}:
\begin{itemize}
 \item \textbf{Replacing the entire $B$ matrix with zeros and then freezing it.} This approach helps us understand the impact of removing the influence of the $B$ matrix from our model, serving as a baseline to assess how essential the $B$ matrix is to the model's performance. The experimental results indicate that removing the influence of the $B$ matrix (by setting it to zero and freezing it) significantly reduced the model's performance across all evaluation metrics. This confirms the crucial role of the $B$ matrix in maintaining baseline model performance, demonstrating its indispensability.
 
\item \textbf{Replacing the entire $B$ matrix with ones and then freezing it.} This test evaluates the effect of uniform values in the $B$ matrix, which contrasts the zero matrix scenario by providing a constant, non-zero bias to the activations. This outcome reflects the impact of uniform non-zero values in the $B$ matrix on model activations, which were ineffective in enhancing performance. The results suggest that a constant, uniform activation value does not facilitate effective learning and adaptability in the model.
 
\item \textbf{Randomly initializing the \(B\) matrix and then freezing it.} This approach was adopted to rule out the possibility that any randomly initialized \(B\) matrix might replicate the performance enhancements observed with our extracted \(B\) matrix. Models employing a randomly initialized and frozen \(B\) matrix demonstrated superior performance compared to those with matrices initialized entirely with zeros or ones. This was noticeable in the yahma\_48\_ran and code\_48\_ran configurations, which showed improved performance compared to the other two treatments. These findings support our hypothesis that appropriately initialized \(B\) matrices could enhance overall model performance and confirm that our extracted common basis matrix is a favorable replacement strategy.
\end{itemize}

The experimental results suggest a noteworthy influence of the $B$ matrix within the model and affirm that proper initialization of the $B$ matrix is key to enhancing model performance. Furthermore, the experiments with randomly initialized $B$ matrices indicate that the selection and optimization process of the $B$ matrix is crucial for achieving optimal model performance.
\section{Analysis}

To objectively and impartially evaluate the performance of the Generation model, we employed GPT-4 \cite{achiam2023gpt} to score the model's outputs in conjunction with the original dataset results. Our assessment focuses on three aspects: Fluency, Relevance, and Accuracy \cite{yang2024gpt}. We aim to determine whether the model's outputs are semantically fluent, whether the content aligns with human interpretative expectations, and how relevant the generated results are to the intended outcomes. Additionally, accuracy is assessed by examining the consistency of the model's outputs with the original data. This set of evaluation criteria is designed to ensure that the model's outputs meet our expected standards.

\begin{table}
\centering
\scalebox{0.8}{
\begin{tabular}{lccc}
\toprule
\textbf{Model} & \textbf{Average Fluency} & \textbf{Average Relevance} & \textbf{Average Accuracy} \\
\midrule
CoRA\_code\_FB\_5 & 87.53 & 76.26 & 79.67 \\
CoRA\_code\_TB\_5 & 88.90 & 81.91 & 82.79 \\
CoRA\_yahma\_TB\_5 & 90.39 & 83.90 & 83.92 \\
CoRA\_yahma\_FB\_5 & 87.79 & 83.35 & 83.49 \\
\bottomrule
\end{tabular}
}
\caption{Scoring Results: Evaluating model outputs' fluency, relevance, and accuracy using GPT-4.}
\label{GPT-4}
\end{table}

The specific results are in the table \ref{GPT-4}. The objective evaluation of GPT-4 shows that all four models achieved the expected scores. This demonstrates that our approach fulfilled the experiment's objectives and ensured the utility of our Generation model's outputs.

Additionally, utilizing a text-to-code dataset, we manually reviewed and executed the generated code to assess its readability and correctness. The scoring outcomes are detailed in Table \ref{human} as follows:

\begin{table}
\centering
\scalebox{0.8}{
\begin{tabular}{lcc}
\toprule
\textbf{Model} & \textbf{Average Readability} & \textbf{Average Correctness}  \\
\midrule
CoRA\_code\_TB\_5 & 87.72 & 90.30  \\
CoRA\_code\_FB\_5 & 87.65 & 90.43 \\
\bottomrule
\end{tabular}
}
\caption{Evaluation Results: Manual inspection of code with scoring based on readability and accuracy.}
\label{human}
\end{table}

Upon evaluating and summarizing, we find that the code's clarity and ease of understanding are high. This indicates that the generated code has a logical structure, follows naming conventions, and is well-commented, allowing our reviewers to comprehend the logic and functionality of the code easily.

The high accuracy score signifies that the code achieves the intended functions and closely matches the specification requirements in execution. This generally means that most of the generated code performs stably across various test scenarios, has a low error rate, and can reliably complete the tasks it was designed to perform.

\section{Conclusion}

We present two innovative optimization schemes for LoRA strategies, successfully demonstrating how to conserve computational resources without sacrificing the performance of large language models. By substituting matrix $B$, our method reduces the number of parameters during model training, maintaining the effectiveness of fine-tuning and, in some cases, surpassing the performance of traditional LoRA parameters. Additionally, using this common basis matrix as an optimized initial state for matrix $B$ training has shown favorable results compared to traditional LoRA. These achievements emphasize the importance of considering computational efficiency in model design and optimization and provide viable strategies for training large models in resource-constrained environments. In the future, we may explore whether there are more optimal structures for employing the common basis matrix. This could further reduce parameters while better-supporting model transfer and participation in downstream task training.
\clearpage

\appendix
\bibliography{CoRA/CoRA}

\end{document}